\documentclass[review,12pt]{elsarticle}

\usepackage{amssymb}
\usepackage{amsmath}
\usepackage{amsmath,amssymb,amsfonts}
\usepackage{algorithmic}
\usepackage{subcaption}
\usepackage{graphicx}
\usepackage{multirow}
\usepackage[table,xcdraw]{xcolor}
\usepackage{xcolor}
\usepackage{rotating}
\usepackage{lscape}

\usepackage{lineno}

\journal{J. of the Mechanical Behavior of Biomedical Materials}

\begin{document}

\begin{frontmatter}



\title{Dual-Variable Force Characterisation method for Human-Robot Interaction in Wearable Robotics}

\author[label1,label3]{Felipe Ballen-Moreno} %% Author name

\affiliation[label1]{organization={Brubotics, Federal Labs AI and Robotics, Vrije Universiteit Brussel},%Department and Organization
            addressline={Pleinlaan 2}, 
            city={Brussels},
            postcode={1050}, 
            country={Belgium}}
\affiliation[label2]{organization={IMEC},%Department and Organization
            addressline={Kapeldreef 75}, 
            city={Leuven},
            postcode={3001}, 
            country={Belgium}}
\affiliation[label3]{organization={Flanders Make},%Department and Organization
            city={Brussels},
            postcode={1050}, 
            country={Belgium}}

\author[label1,label2]{Pasquale Ferrentino}
\author[label1,label2]{Milan Amighi}
\author[label1,label2]{Bram Vanderborght}
\author[label1,label3]{Tom Verstraten}


%% Abstract
\begin{abstract}
Understanding the physical interaction with wearable robots is essential to ensure safety and comfort. However, this interaction is complex in two key aspects: (1) the motion involved, and (2) the non-linear behaviour of soft tissues. Multiple approaches have been undertaken to better understand this interaction and to improve the quantitative metrics of physical interfaces or cuffs. As these two topics are closely interrelated, finite modelling and soft tissue characterisation offer valuable insights into pressure distribution and shear stress induced by the cuff. Nevertheless, current characterisation methods typically rely on a single fitting variable along one degree of freedom, which limits their applicability, given that interactions with wearable robots often involve multiple degrees of freedom. To address this limitation, this work introduces a dual-variable characterisation method, involving normal and tangential forces, aimed at identifying reliable material parameters and evaluating the impact of single-variable fitting on force and torque responses. This method demonstrates the importance of incorporating two variables into the characterisation process by analysing the normalized mean square error (NMSE) across different scenarios and material models, providing a foundation for simulation at the closest possible level, with a focus on the cuff and the human limb involved in the physical interaction between the user and the wearable robot.

\end{abstract}

\begin{graphicalabstract}
\includegraphics[width=\textwidth]{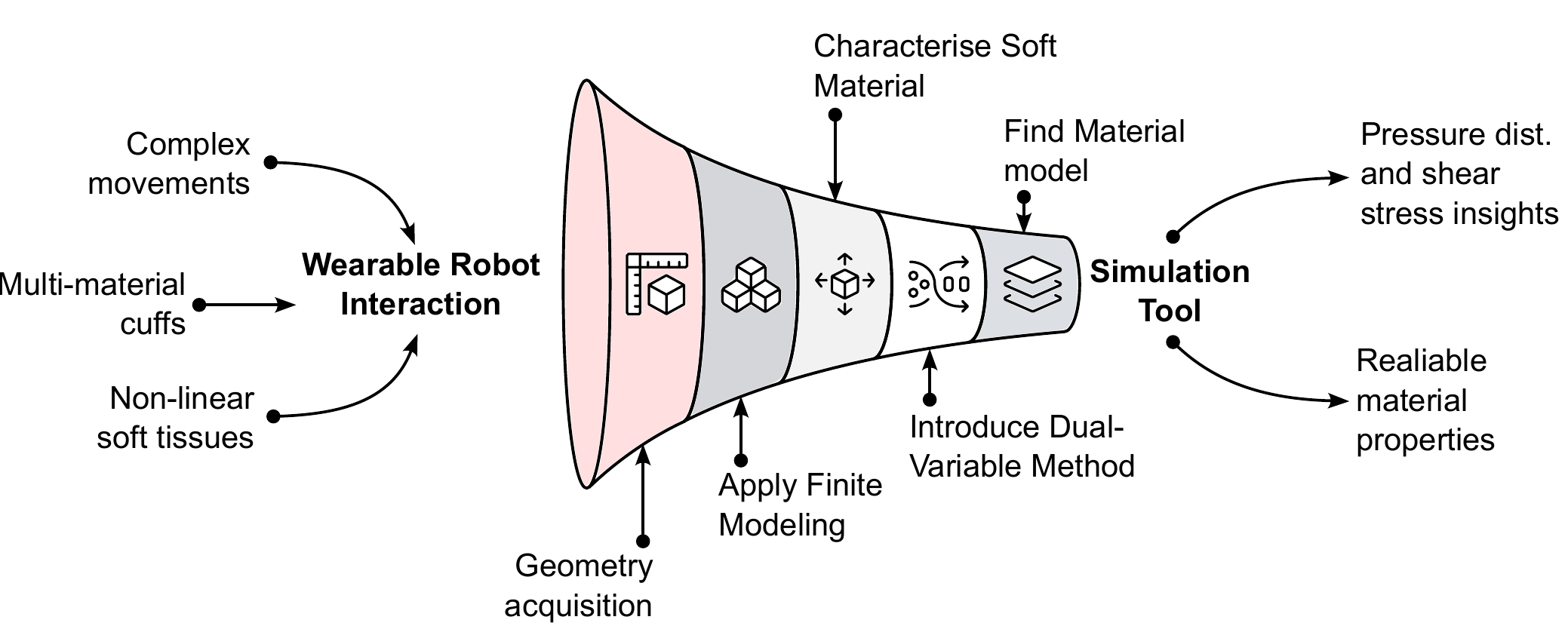}
\end{graphicalabstract}

\begin{highlights}
\item The resulting force or torque responses deviate from the experimental data when the characterisation process only considers a single variable during curve fitting and parameter estimation.
\item Two variables into the indentation characterisation allows to find a better set of parameters of a material model that interacts and provides normal and shear forces in a realistic way.
\end{highlights}

%% Keywords

\end{frontmatter}

% \linenumbers


\section{Introduction}

One of the main components of wearable robotics are the physical interfaces or cuffs, which attach the device to the human. The design of physical interfaces or cuffs  depends on the final application, with interactions evaluated through qualitative and quantitative metrics such as force, torque, pressure, relative motion, and user feedback \cite{Massardi2022}. Quantitative assessments often rely on experimental data or anatomical models to objectively capture the skin and soft tissue response to specific loads \cite{Lenzi2011}.

To reduce the reliance on extensive experimentation and the proliferation of prototypes, multiple approaches have been proposed to understand the physical interactions between the interface and the human body \cite{Massardi2022,Pinto-Fernandez2020}. These approaches are divided into: sensorised cuffs or external equipment, phantom or artificial limbs, and soft tissue characterisation processes. Multiple sensors or equipment within the device or its surroundings to capture and analyse specific tasks \cite{Pinto-Fernandez2020}. This method yields localized (e.g., force or pressure sensors) or indirect (e.g., motion capture data) information about the interaction. For instance, force-sensing resistors (FSRs) are widely implemented as matrices within cuffs to capture regular forces \cite{Bessler-Etten2024,Turnbull2023,Yousaf2021}. Similarly, mechano-optoelectronic sensors are embedded to measure normal and tangential forces or estimate pressure resulting from the interaction \cite{Georgarakis2018,Lenzi2011}. Nevertheless, these sensors are limited in providing localised information, which depends on both the number and placement of the sensors within the cuff. 

Data from localised sensors are insufficient to assess the interaction due to the limited information regarding pressure distribution, shear stress behaviour resulting from complex motions, or slippage between the skin and the cuff. These phenomena are understood as the internal dynamics between the skin and the physical interface. Furthermore, such data are highly dependent on the internal positioning of bones and the dominant underlying tissues, including muscles, tendons, ligaments, and the thickness of the adipose layer. Therefore, wide-range and direct approaches hold promise for quantifying the interaction between the skin and the soft tissue response to specific loads and cuff configurations  \cite{Lenzi2011}. The connection among the cuff, skin, and soft tissue is commonly described through stiffness and/or damping components \cite{JohnVarghese2022,Langlois2021b,Yousaf2021}. However, models based on these components may not provide direct predictions on the distribution of normal and shear stresses or pressure caused at the surface level of the interaction, due to their general definition or boundary conditions between the components \cite{Ballit2022}. 

The construction of artificial limbs, known as phantoms, mimics human anatomy to a certain degree, thereby simplifying the physical interaction and isolating specific outcomes of interest. However, these phantoms exhibit limitations in replicating human limb and joint geometry and material properties \cite{Barrutia2024,Deman2023}. To characterise internal dynamics at the interface, soft tissue models have been developed to investigate the stress and pressure distribution caused by the cuff, typically characterising a limb area across one degree of freedom using a material model which is one of their main limitations. Non-linear material models are commonly defined to replicate the soft tissue complex, including geometry, through medical imaging techniques (e.g., superficial 3D scans, CT, or MRI), where one or more anatomical structures (e.g., bones, skin, muscles, tendons, ligaments, adipose layer) are incorporated into the characterisation process \cite{Frauziols2016,Portnoy2008,Ranger2023,Sengeh2016,Tran2007,Arnold2023}. As more structures are included, the number of model parameters increases, along with the complexity of matching experimental data.

Characterisation processes can involve indentation, shear wave elastography, and ultrasound data to fit the non-linear model to experimental results  \cite{Ranger2023,Sengeh2016}. Current methodologies typically focus on a single variable (or characterisation degree of freedom, understood as one deformation mode) for fitting, due to the complexity and non-linearities of the soft tissues. In the case of indentation, this is the linear displacement applied to the tissue (i.e., deformation mode: uniaxial compression); shear wave elastography uses the propagation speed of the wave through the soft tissue (i.e., uniaxial tension and compression in the orthogonal direction or pure shear). The selected variable enables the identification of a parameter set that fits the experimental response. However, relying on a single variable to characterise soft tissue may limit the model’s ability to predict interactions not aligned with that variable and may result in multiple parameter sets yielding similar fits. Oddes et al. demonstrated that numerous solutions can fall within the same 7–10\% error range when only indentation force is considered \cite{Oddes2023}. 

As previous methods rely on a single variable, one single deformation mode, and provide a plurality of parameter sets, this work proposes a methodology that incorporates two variables to characterise soft materials by identifying a unique set of model parameters. In this sense, the response of the soft material will be characterised by two representative deformation modes: uniaxial compression and shear. In addition, our approach replicates the boundary conditions imposed along a forearm phantom by bones, the geometry involved in the characterisation region, and the variation in material properties resulting from the inclusion of a second variable in the characterisation process.

This article is structured as follows. Section II presents how the phantom was made, how the indentation was performed, then, a description of material models used to fit the experimental data, the finite element model on FEBio, the parameter estimation using Latin hypercube sampling, and finally the data processing of the force and torque experimental curves. Section III presents the fitting results for each material model and the analysis on the parameters set when one or both experimental curves are include in the fitting process. Section IV discusses the findings and implications, and Section V concludes with future work.

\section{Methods}

This section describes the fabrication of the phantom and the indentation protocol, followed by a description of the material models used in the fitting process, the in silico scenarios on FEBio software, the parameter estimation process using Latin hypercube sampling, and finally the data processing of the force and torque experimental curves.

\subsection{Forearm phantom}

\begin{figure}
    \centering
    \includegraphics{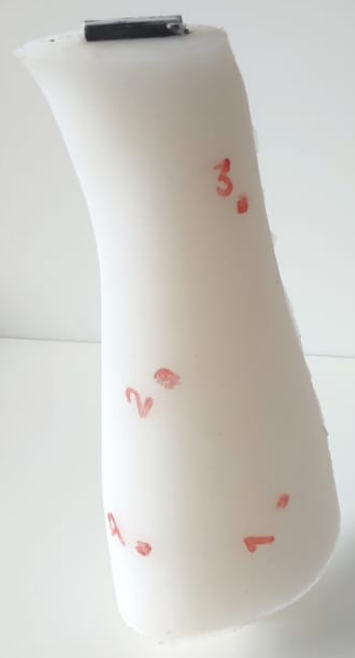}
    \caption{Forearm phantom. The phantom is composed of Ecoflex 0050. The indentation spots are marked with red ink, providing positional references for subsequent simulations. Each spot represents different boundary conditions as the representative bones and the Ecoflex volume involved in the indentation vary.}
    \label{fig:phantom}
\end{figure}

The phantom was created using two steps: (1) a 3D surface scan (Structure Sensor, SF, USA) of a healthy male subject (22 y.o., 1.77 m, 74 kg), and (2) a PLA 3D-printed representation of the radius and ulna bones from \cite{Hamze2020}. A mold was designed to replicate the scanned surface and to fix the bones in an anatomically accurate position. Ecoflex 0050 (Smooth-on, PA, USA) was used to represent the hyperelastic behavior of soft tissue, pouring a 1:1 ratio mixture into the mold, given the forearm phantom shown in Fig. \ref{fig:phantom}. These materials were chosen because their Young's Modulus are of a similar order of magnitude.

\subsection{Experimental tests}

Experimental tests were conducted at four points on the forearm phantom using a spherical indenter (diameter 15 mm) with an normal indentation depth of 10 mm. The motion was defined by a cosine function, where the amplitude corresponds to the indentation depth, followed by a twist ranging from 22.5° to -22.5°. The phantom was supported on a table while the robotic arm (Panda, Franka Robotics, Munich, Germany) performed the indentation. The robot’s end-effector was equipped with a 6-axis force-torque sensor (Medusa, Bota Systems, Zürich, Switzerland) to acquire force and torque interactions. Each indentation point was marked with ink and subsequently 3D scanned to capture the geometry for further simulations.

\subsection{Constitutive models}

Hyperelastic models were used to fit indenting and twisting curves, which represent force-displacement and torque-rotation, respectively. Following the formulation of FEBio, the deformation gradient \textbf{F} and its deviatoric part $\tilde{\textbf{F}}=J^{1/3}\textbf{F}; J=det\textbf{F}$, can be defined as the right Cauchy-Green tensor $\textbf{C}=\textbf{F}^{T}\textbf{F}$ divided into volumetric and deviatoric $\tilde{\textbf{C}}=\tilde{\textbf{F}}^{T}\tilde{\textbf{F}}=J^{-2/3}\textbf{C}$. The nearly-imcompressible condition must satisfy $det(\tilde{\textbf{F}})=1$ and can be simplified by an uncoupled formulation of the strain energy density $\Psi(\textbf{C})$ into a volumetric energy component $U(J)$ and a distortional component $\tilde{\Psi}(\textbf{C})$. Further details on this topic can be refer in \cite{holzapfel2002nonlinear}.
This constitutive behaviour is in function of the invariants of $\textbf{C}$ as $I_1=tr\textbf{C}=\textbf{I},I_2 = \frac{1}{2}[(tr\textbf{C})^2-tr\textbf{C}^2], I_3 = det(\textbf{C})=J^2$. Resulting on the second Piola-Kirchhof stress tensor as:
\begin{equation}
    \textbf{S} = 2\frac{\partial\Psi}{\partial\textbf{C}} = 2\frac{\partial\Psi}{\partial I_1}\frac{\partial I_1}{\partial\textbf{C}} + 2\frac{\partial\Psi}{\partial I_2}\frac{\partial I_2}{\partial\textbf{C}} + 2\frac{\partial\Psi}{\partial I_3}\frac{\partial I_3}{\partial\textbf{C}} \nonumber
\end{equation}

The data was fitted for three different hyperelastic models commonly used to represent soft tissues. The first material model implements the Ogden model following its uncoupled formulation, the first order was used to fit two main parameters $c_1$, $m_1$, shown in Eq. \ref{eq:ogden}.

\begin{equation}
    \Psi = \sum_{i=1}^{N} \frac{c_i}{m^{2}_i}(\tilde{\lambda}^{m_i}_1 +\tilde{\lambda}^{m_i}_2 +\tilde{\lambda}^{m_i}_3 -3) + U(J)
    \label{eq:ogden}
\end{equation}

Second material model uses the third order Yeoh model is chosen where $c_1$, $c_2$, and $c_3$ are the parameters that defined its behaviour. 

\begin{equation}
    \Psi = \sum^3_{i=1} c_i(I_i-3)^i
    \label{eq:yeoh}
\end{equation}

The last material model uses the Neo-Hookean model where the $\mu$ and $\lambda$ are the parameters shown in Eq. \ref{eq:neoh}, that can be translate to the Young modulus $E$ and poisson ratio $v$ as shown in Eq. \ref{eq:Ev_eq}.   

\begin{equation}
    \Psi = \frac{\mu}{2}(I_1 -3) - \mu lnJ + \frac{\lambda}{2}(lnJ)^2
    \label{eq:neoh}
\end{equation}
\begin{equation}
    \mu = \frac{E}{2(1+v)} ; \lambda = \frac{vE}{(1+v)(1-2v)}\\
    \label{eq:Ev_eq}
\end{equation}

\subsection{Finite element model}

\begin{figure}
    \centering
    \includegraphics{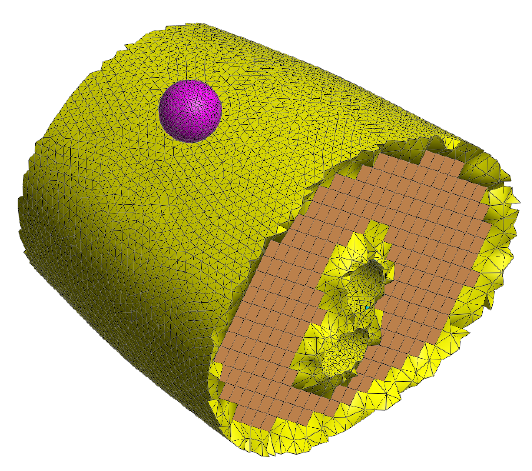}
    \caption{Phantom mesh. A segmentation example illustrates the mesh which two element types are used to model the forearm segment: 45,184 tetrahedral elements represent the superficial geometry, bone boundaries, and the transition between element types; and 3,429 hexahedral elements define the soft tissue complex. For both meshes, the indenter is defined as a rigid body and 1,280 tetrahedral elements.}
    \label{fig:mesh_one}
\end{figure}

The geometry was created using the CAD model of the mold and bone meshes, segmenting the region of analysis based on the colour mesh that marks the indentation location to reduce the number of elements. According to the characterisation spot, the segmented part has a longitudinal length of 80mm. The geometry was re-meshed and volumetrically filled using the MATLAB Toolbox GIBBON \cite{MMoerman2018}. The mesh of the phantom segment includes tetrahedral and hexahedral elements to represent the overall silicone volume. The indenter is represented by a rigid sphere and a tetrahedral mesh, as shown in Fig. \ref{fig:mesh_one}. The tetrahedral elements are chosen to follow complex geometries as the surface mold and the bones, and hexahedral elements to evenly fill the volume. Both element types are set with the same material model.

The boundary conditions fully constrained displacement and rotation at both ends of the deformable geometry, including the bone boundaries as fixed geometry. The sphere was prescribed with rigid displacements and rotations to replicate the indenting and twisting motion described in Section II-A. The contact between the indenter and the segment considered the sliding-elastic behaviour, penalty definition and friction contact (coefficient of friction = 1.0) between PLA 3D-printed material and the phantom silicone.

\subsection{Data processing and FEBio software}

Data analysis for soft tissue characterisation is based on robot displacement and end-effector rotation, which are used to derive force and torque curves from the 6-axis sensor. These curves and in-silico data are interpolated to match displacement and rotation with the corresponding force and torque values. Data processing was carried out in MATLAB R2021b, and simulations were performed using FEBio 4.6 \cite{Maas2012}. The implicit solver was configured with minimal artificial damping. The parametric sweep was executed using the Parallel Computation toolbox of MATLAB on a 24-core cluster (VUB-HCP, Tier-2), running 140 or 250 simulations per characterisation spot and material model. Three scenarios are defined to compare the influence on the material coefficients according to the source of error: (I) the sum of both NMSE , (II) the minimum NMSE considering only the torque and (III) the minimum NMSE considering only the force.

\subsection{Parameters estimation}

To determine the model parameters,the force–displacement curve is fitted to estimate the soft tissue complex material parameters. This is accomplished by minimizing the normalized mean squared error (NMSE) of the indenter force, as shown in Eq. \ref{eq:force_minima}.

\begin{equation}
    Fopt_f = \frac{1}{N_f} \sum \frac{|F_{sim} - F_{exp}|^2}{|\overline{F_{exp}}|^2}
    \label{eq:force_minima}
\end{equation}

Where $F_{sim}$ is the force of the simulation, $F_{exp}$ is the experimental force, and $N_f$ is the number of points included to estimate the NMSE. Similarly, Eq. \ref{eq:force_minima} is also used to estimate the NMSE of the torque-rotation curves. Finally, both the force–displacement and torque–rotation curves to estimate the overall NMSE:

\begin{equation}
    Fopt =  \frac{1}{N_f} \sum\frac{|F_{sim} - F_{exp}|^2}{|\overline{F_{exp}}|^2} +  \frac{1}{N_t} \sum\frac{|T_{sim} - T_{exp}|^2}{|\overline{T_{exp}}|^2}
    \label{eq:tor-for_minima}
\end{equation}

Where $T_{sim}$ is the simulation torque, $T_{exp}$ is the experimental torque, and $N_t$ is the number of points included to estimate the NMSE.

The region of interest per model is defined as follows: for the Ogden model, $m_1 = [1,8], c_1 = [3e^{-2}, 2e^{-1}], \kappa = [2.5e^{-1}, 2.5]$; for the Yeoh model, $c_1 = [1.4e^{-3}, 3e^{-2}], c_2 = [-4.14e^{-5}, -3e^{-3}], c_3 = [3e^{-6}, 3e^{-4}]$; and for the Neo-Hookean model, $E = [1e^{-3}, 1], v = [0.40, 0.49]$. Each region defines plausible ranges within which the force and torque magnitudes are reachable. The parameter sweeper identifies and evaluates suitable fits to the model at the lowest NMSE, as shown in Eq. \ref{eq:tor-for_minima}. Parameter selection was performed using 250 sets generated by Latin hypercube sampling from the Design of Experiments toolbox in MATLAB R2021b. The Drucker criterion was considered to ensure a stable parameter set, reducing to 140 sets for the Yeoh model. 

\subsection{Generalisation of parameters}

According to the material model that presented the lower NMSE, each of the experimental data was fitted to the corresponding in silico force and torque, finding a set of parameters through the lowest sum of NMSE. In this sense, the mean of NMSE was calculated, as shown in Eq. \ref{eq:global_tor-for}.

\begin{equation}
    Fopt_G = \frac{1}{4}\sum_{i=1}^{4} Fopt_{ij} 
    \label{eq:global_tor-for}
\end{equation}

Where $i$ is the experimental point, $j$ the number of the set of parameters with the lowest sum of NMSE, and $Fopt$ is the error defined in Eq. \ref{eq:tor-for_minima}. 

\section{Results}

This section presents the material model analysis based on both experimental and in-silico results. The analysis is divided into two main approaches: (1) experimental data from each characterisation spot, and (2) the importance of including torque–rotation data in the characterisation process across three material models.

\subsection{Two-variable characterisation}

\begin{figure}
    \centering
    \includegraphics[scale=1]{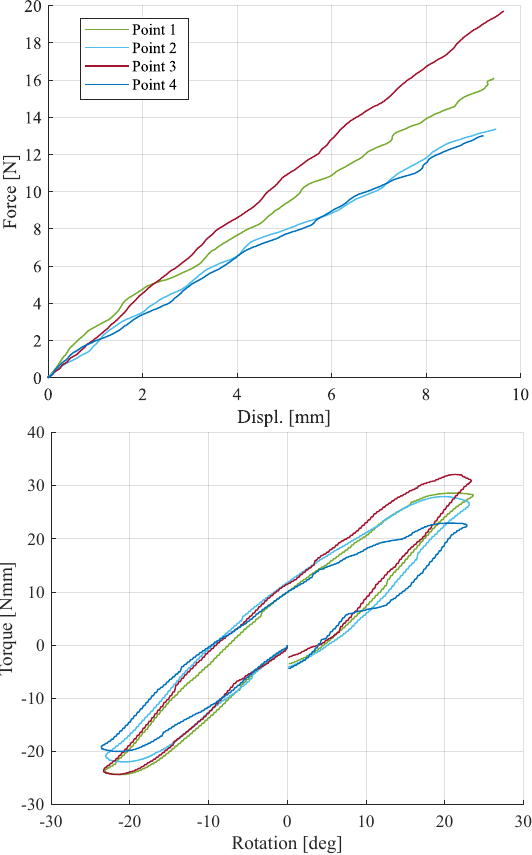}
    \caption{Experimental results of the forearm phantom at four points. Force and torque data are illustrated with continuous lines. The force-displacement curves are shown at the top, and the torque-rotation curves are at the bottom.}
    \label{fig:exp_data}
\end{figure}

The experimental data are divided into force–displacement and torque –rotation curves per characterisation spot, as shown in Fig. \ref{fig:exp_data}. The force –displacement curves are shown at the top of Fig. \ref{fig:exp_data}, and the torque–rotation curves at the bottom. Each spot exhibited a slightly different response, with maximum indentation forces ranging from 13 to 19.6 N, and torque responses between –24 and 32 Nmm. As the phantom has a uniform distribution of silicone, the response ranges for force and torque differ per point due to the boundary conditions, which create this slightly different response.

\subsection{Fitting process across different models}

Including torque response in the characterisation process influenced the curve fitting depending on the selected material model, as the behavior may better resemble either force or torque. The following analysis presents the NMSE behavior across the parameter space, showing force–displacement and torque–rotation fits for three scenarios: (I) the sum of both NMSE values (red dotted curves), (II) the minimum NMSE considering only torque (yellow dotted curves), and (III) the minimum NMSE considering only force (purple dotted curves). Finally, the results of the summed NMSE per parameter set are shown, as each model exhibits a minimum threshold.

\subsubsection{Ogden first order}

\begin{figure}
    \centering
    \includegraphics{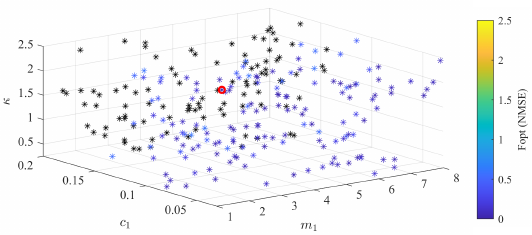}
    \caption{Ogden coefficient space. It illustrates 250 sets of coefficients and their corresponding NMSE values, ranging from 0 to 2.5 (color scale). Points outside this range are shown in black, and the minimum NMSE is highlighted in red. Point 3 is shown to exemplify the behaviour of the NMSE in the parameter space. Distribution of NMSE across 250 parameter sets per point, illustrating a minimum at 0.1034 and varying NMSE ranges per point. The upper limit is set at 4.0 to highlight the threshold and range, as a few sets exceed this value.}
    \label{fig:ogden_errors}
\end{figure}

The parameter sweep for point 3 reveals a distinct NMSE distribution ranging from 0 to 2.5, as shown in Fig. \ref{fig:ogden_errors}. Each analysis point displays a similar NMSE trend. For instance, points 1 and 3 shows three zones delimited in black, light and dark blue clusters along the $c_1$ axis. Points 2 and 4 exhibit two clusters in black and dark blue. The coefficient sets with the lowest NMSE for points 1, 2, and 3 are more consistent than those for point 4, which contains a greater number of coefficient pairs with NMSE values exceeding 2.5, shown in black.

\begin{figure}
    \centering
    \includegraphics{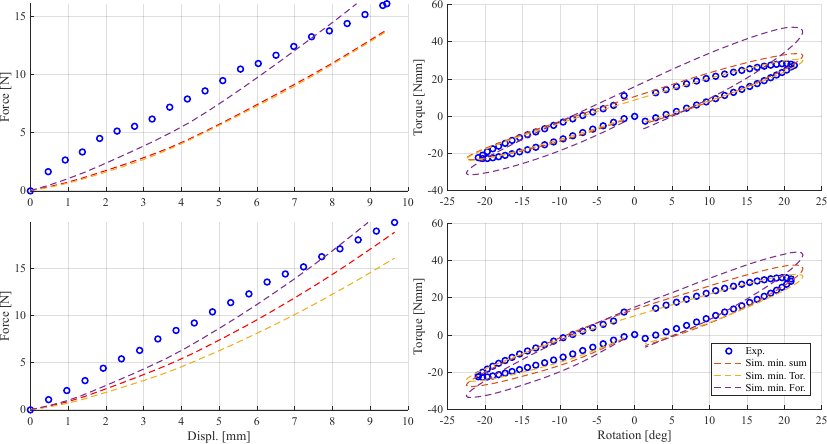}
    \caption{Force and torque responses for the Ogden model. Dotted lines represent three fitted curves based on the minimum NMSE of the sum of force and torque errors (red curves), the minimum NMSE considering only torque error (yellow curves), and the minimum NMSE considering only force error (purple curves). Blue circles indicate experimental data. Each row illustrates the response for point 1 and 3, top to bottom. Two dotted lines, shown in the top-left figure, as the yellow and red curves are identical. Second row represents the response for point 3.}
    \label{fig:ogden_kinet}
\end{figure}

Force and torque curves fit differently depending on the NMSE criteria used in the characterisation process, as shown in Fig. \ref{fig:ogden_kinet}, considering point 1 and point 3 into the following analysis. Red curves closely matches the experimental force shown in blue; however, three out of four points exhibit higher torque values. Similarly, the yellow curves align more closely with the experimental torque data, although the force curves deviate from the experimental data. Point 1 shows alignment between the yellow and red curves. Similarly, point 2 also shows alignment between Sim. min. Tor. and Sim. min. sum curves. In this sense, the identification of the model parameters is feasible by including force and torque curves. 

\subsubsection{Yeoh third order}

\begin{figure}
    \centering
    \includegraphics{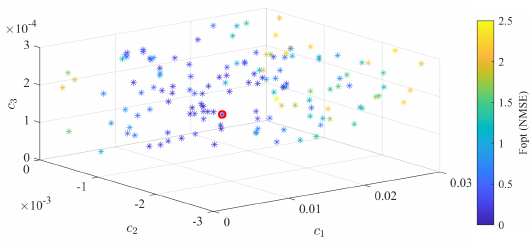}
    \caption{Yeoh coefficient space. It illustrates 140 sets of coefficients and their corresponding NMSE values, ranging from 0 to 2.5 (color scale). Points outside this range are shown in black, and the minimum NMSE is highlighted in red. In this case point 2 is used to exemplify the behaviour of NMSE. The upper limit is set at 4.0 to highlight the threshold and range, as a few sets exceed this value.}
    \label{fig:yeoh_coeffs}
\end{figure}

The parameter space of this model is presented in Fig. \ref{fig:yeoh_coeffs}, showing the stable set of coefficients based on the Drucker criteria. Each point exhibits a similar NMSE distribution ranging from 0 to 2.5. The analysis reveals consistent trends across points; for instance, all points show elevated NMSE values at higher $c_1$, however, point 1 also shows few sets with a higher NMSE for the lowest $c_1$. Points 2 (Fig. \ref{fig:yeoh_coeffs}) and 4 display clustered NMSE values, with color scales between 0.5 and 1.5. The coefficient sets with the lowest NMSE for each point are between 0.01-0.02 for $c_1$, nevertheless, the $c_1$ and $c_3$ ranges less consistent. The greater number of coefficient pairs with NMSE values exceeding 2.5.

\begin{figure}
    \centering
    \includegraphics{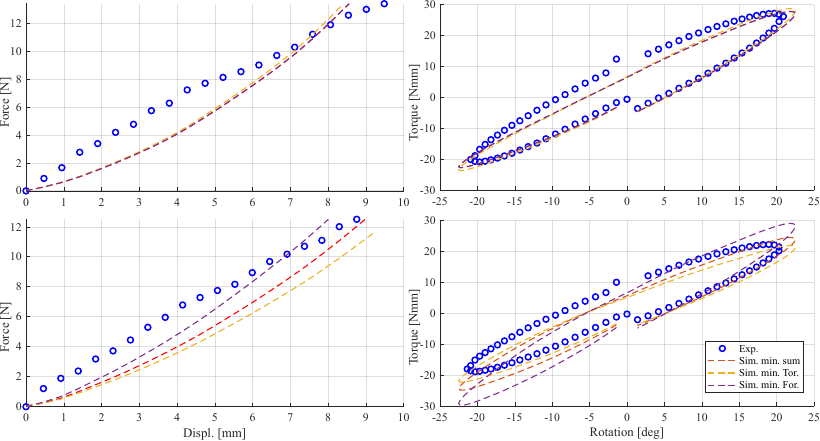}
    \caption{Force and torque responses for the Yeoh model. Dotted lines represent three fitted curves based on the minimum NMSE of the sum of force and torque errors (red curves), the minimum NMSE considering only torque error (yellow curves), and the minimum NMSE considering only force error (purple curves). Blue circles indicate experimental data. Each row illustrates the response for point 2 and 4, top to bottom. Two dotted lines, shown in the top row, as the purple and red curves are identical.}
    \label{fig:yeoh_kinet}
\end{figure}

Force and torque curves fit differently depending on the NMSE criteria used in the characterisation process, as shown in Fig. \ref{fig:yeoh_kinet}. For this model, Fig. \ref{fig:yeoh_kinet} display the results of point 2 and 4. The purple curves provide a better fit to the experimental force curves for each point; however, they also exhibit higher NMSE values and magnitudes for the torque response. The yellow curves deviate further from the experimental torque data, as shown in bottom-right Fig. \ref{fig:yeoh_kinet}. Additionally, the force curves in this scenario diverge from the experimental data. Point 1 shows alignment between the purple andred curves, as illustrated in the first row Fig. \ref{fig:yeoh_kinet}. In this context, including two variables into the fitting process provide a better response along the normal force and shear direction.

\subsubsection{Neo-Hookean}

\begin{figure}
    \centering
    \includegraphics{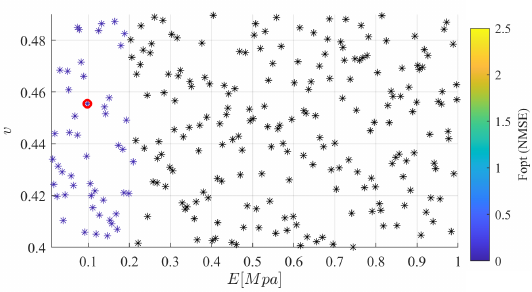}
    \caption{Neo-Hookean coefficient space. It illustrates 250 sets of coefficients and their corresponding NMSE values, ranging from 0 to 2.5 (color scale). Points outside this range are shown in black, and the minimum NMSE is highlighted in red. Point 3 is used to show the NMSE behaviour for this model. Distribution of NMSE across 250 parameter sets per point, illustrating a wide spread of NMSE values. The upper limit is set at 4.0 to highlight the scatter behaviour, although a few sets exceed this threshold.}
    \label{fig:neoh_coeffs}
\end{figure}

The parameter sweeper for each point reveals a clear NMSE distribution up to 0.2 MPa, as shown in Fig. \ref{fig:neoh_coeffs}. All four analysis points display similar NMSE trends. Points 1, 2, and 4 share a region with minimum NMSE values, whereas point 3 exhibits lower $v$ and $E$ values. Additionally, point 3 contains a greater number of coefficient pairs with NMSE values exceeding 2.5, shown in black.

\begin{figure}
    \centering
    \includegraphics{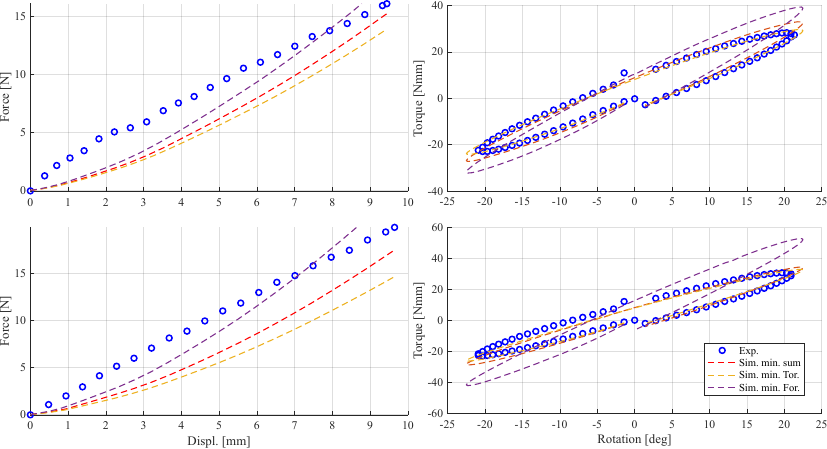}
    \caption{Force and torque responses for the Neo-Hookean model. Dotted lines represent three fitted curves based on the minimum NMSE of the sum of force and torque errors (red curves), the minimum NMSE considering only torque error (yellow curves), and the minimum NMSE considering only force error (purple curves). Blue circles indicate experimental data. Each row illustrates the response for point 1 and 3, top to bottom.}
    \label{fig:neoh_kinet}
\end{figure}

Force and torque curves fit differently depending on the NMSE criteria used in the characterisation process, as shown in Fig. \ref{fig:neoh_kinet}. The red curves provide a better fit to the experimental force curves at each point compared to the Ogden model and yield similar NMSE values to the Yeoh model. However, they also exhibit higher NMSE values and magnitudes for the torque response. The yellow curves deviate further from the experimental torque data for point 2 and 4. Additionally, the force curves in this scenario diverge from the experimental data. Point 1 shows alignment between the yellow and red curves, as illustrated in first row of Fig. \ref{fig:neoh_kinet}. Following the same tendency of previous models, including two variables into the fitting process provide a parameter set that response closely to the experimental data for both directions, normal and shear.

\subsection{Set of coefficients on multiple points}

The NMSE behaves differently across material models, scenarios, and characterisation spots, as shown in Table \ref{tab:results_coeffs}. Firstly, the set of coefficients per scenario varies considerably for each material model, except in cases where the minimum NMSE corresponds to scenario II. Additionally, these coefficients cluster within a specific region in three out of four points for the Ogden and Neo-Hookean models. 

% \begin{landscape}
    \begin{table}
        % \centering
        \caption{Set of coefficients per scenario. Each scenario shows (I) the sum of both NMSE, (II) the minimum NMSE considering only the torque, and (III) the minimum NMSE considering only the force. Loc. stands for location point of analysis.}
        \label{tab:results_coeffs}
        \resizebox{0.70\textwidth}{!}{\begin{minipage}{\textwidth}
        \begin{tabular}{cccccccccc}
        \hline
        \multicolumn{10}{c}{Ogden first order} \\
        \multicolumn{1}{c|}{\multirow{2}{*}{Loc.}} & \multicolumn{3}{c|}{\textbf{I}} & \multicolumn{3}{c|}{\textbf{II}} & \multicolumn{3}{c}{\textbf{III}} \\
        \multicolumn{1}{c|}{} & \textbf{$m_1$} & \textbf{$c_1$} & \multicolumn{1}{c|}{\textbf{$\kappa$}} & \textbf{$m_1$} & \textbf{$c_1$} & \multicolumn{1}{c|}{\textbf{$\kappa$}} & \textbf{$m_1$} & \textbf{$c_1$} & \multicolumn{1}{c}{\textbf{$\kappa$}} \\
        \multicolumn{1}{c|}{1} & 1.9152 & 0.0590 & \multicolumn{1}{c|}{1.0488} & 1.0127 & 0.0608 & \multicolumn{1}{c|}{0.4826} & 2.4794 & 0.0741 & \multicolumn{1}{c}{2.4150} \\
        \multicolumn{1}{c|}{2} & 2.3832 & 0.0464 & \multicolumn{1}{c|}{0.8192} & 4.9542 & 0.0313 & \multicolumn{1}{c|}{1.1841} & 2.3832 & 0.0464 & \multicolumn{1}{c}{0.8192} \\
        \multicolumn{1}{c|}{3} & 2.0883 & 0.0598 & \multicolumn{1}{c|}{2.3370} & 2.2875 & 0.0546 & \multicolumn{1}{c|}{1.2777} & 2.7450 & 0.0732 & \multicolumn{1}{c}{2.1741} \\
        \multicolumn{1}{c|}{4} & 6.4097 & 0.0428 & \multicolumn{1}{c|}{2.1606} & 6.6523 & 0.0382 & \multicolumn{1}{c|}{1.9982} & 1.9152 & 0.0590 & \multicolumn{1}{c}{1.0488} \\ \hline
        \multicolumn{10}{c}{Yeoh third order} \\
        \multicolumn{1}{c|}{\multirow{2}{*}{Loc.}} & \multicolumn{3}{c|}{\textbf{I}} & \multicolumn{3}{c|}{\textbf{II}} & \multicolumn{3}{c}{\textbf{III}} \\
        \multicolumn{1}{c|}{} & \textbf{$c_1$} & \textbf{$c_2$} & \multicolumn{1}{c|}{\textbf{$c_3$}} & \textbf{$c_1$} & \textbf{$c_2$} & \multicolumn{1}{c|}{\textbf{$c_3$}} & \textbf{$c_1$} & \textbf{$c_2$} & \multicolumn{1}{c}{\textbf{$c_3$}} \\
        \multicolumn{1}{c|}{1} & 1.4651e-2 & -1.2387e-3 & \multicolumn{1}{c|}{1.2673e-4} & 1.2922e-2 & -2.0169e-3 & \multicolumn{1}{c|}{2.7623e-4} & 1.9089e-2 & -1.7694e-3 & \multicolumn{1}{c}{2.1710e-4} \\
        \multicolumn{1}{c|}{2} & 1.0383e-2 & -7.3184e-4 & \multicolumn{1}{c|}{3.7493e-5} & 1.0666e-2 & -9.4276e-4 & \multicolumn{1}{c|}{2.7784e-4} & 1.0383e-2 & -7.3184e-4 & \multicolumn{1}{c}{3.7493e-5} \\
        \multicolumn{1}{c|}{3} & 1.5858e-2 & -2.1409e-3 & \multicolumn{1}{c|}{2.3770e-4} & 1.3786e-2 & -8.3150e-4 & \multicolumn{1}{c|}{1.2858e-4} & 1.9089e-2 & -1.7694e-3 & \multicolumn{1}{c}{2.1710e-4} \\
        \multicolumn{1}{c|}{4} & 1.2080e-2 & -4.4848e-5 & \multicolumn{1}{c|}{1.6647e-4} & 1.0885e-2 & -4.8000e-4 & \multicolumn{1}{c|}{2.5510e-4} & 1.4651e-2 & -1.2387e-3 & \multicolumn{1}{c}{1.2673e-4} \\ \hline
        \multicolumn{10}{c}{Neo-hookean} \\
        \multicolumn{1}{c|}{\multirow{2}{*}{Loc.}} & \multicolumn{3}{c|}{\textbf{I}} & \multicolumn{3}{c|}{\textbf{II}} & \multicolumn{3}{c}{\textbf{III}} \\
        \multicolumn{1}{c|}{} & \multicolumn{2}{c}{\textbf{$E[Mpa]$}} & \multicolumn{1}{c|}{\textbf{$v$}} & \multicolumn{2}{c}{\textbf{$E[Mpa]$}} & \multicolumn{1}{c|}{\textbf{$v$}} & \multicolumn{2}{c}{\textbf{$E[Mpa]$}} & \multicolumn{1}{c}{\textbf{$v$}} \\
        \multicolumn{1}{c|}{1} & \multicolumn{2}{c}{0.1071} & \multicolumn{1}{c|}{0.4200} & \multicolumn{2}{c}{0.0954} & \multicolumn{1}{c|}{0.4353} & \multicolumn{2}{c}{0.1269} & \multicolumn{1}{c}{0.4125} \\
        \multicolumn{1}{c|}{2} & \multicolumn{2}{c}{0.0735} & \multicolumn{1}{c|}{0.4434} & \multicolumn{2}{c}{0.0735} & \multicolumn{1}{c|}{0.4434} & \multicolumn{2}{c}{0.0735} & \multicolumn{1}{c}{0.4434} \\
        \multicolumn{1}{c|}{3} & \multicolumn{2}{c}{0.0976} & \multicolumn{1}{c|}{0.4555} & \multicolumn{2}{c}{0.0911} & \multicolumn{1}{c|}{0.4058} & \multicolumn{2}{c}{0.1460} & \multicolumn{1}{c}{0.4044} \\
        \multicolumn{1}{c|}{4} & \multicolumn{2}{c}{0.0732} & \multicolumn{1}{c|}{0.4850} & \multicolumn{2}{c}{0.0636} & \multicolumn{1}{c|}{0.4239} & \multicolumn{2}{c}{0.0782} & \multicolumn{1}{c}{0.4842} \\ \hline
        \end{tabular}
        \end{minipage} }
    \end{table}
% \end{landscape}

The overall minimum NMSE found is 0.0853 for the Yeoh model in scenario I at indentation point 2. Nevertheless, this model presents the highest error for the remaining indentation points compared to the other models. The intermediate NMSE of 0.0959 for the Neo-hookean model in scenario I also occurs at point 4, and scenario II achieves the lowest NMSE. The Ogden model shows an NMSE of 0.1034 in scenario I at indentation point 3 and the lowest values in scenarios II and III of 0.0037 and 0.0274, respectively. The difference between the maximum and minimum NMSE also varies per model and scenario summarised in Table \ref{tab:results_errors}, highlighting the lower and higher value per scenario in green and red, respectively.

\begin{table}
    \centering
    \caption{NMSE per scenario and point. Each scenario shows (I) the sum of both NMSE, (II) the minimum NMSE considering only the torque, and (III) the minimum NMSE considering only the force. Values highlighted in green and red relates the lower and higher value per scenario, respectively. Loc. stands for location point of analysis.}
    \begin{tabular}{cccc}
    \hline
    \multicolumn{4}{c}{Ogden first order} \\
    \multicolumn{1}{c|}{} & \multicolumn{3}{c}{NMSE} \\
    \multicolumn{1}{c|}{\multirow{-2}{*}{Loc.}} & \textbf{I} & \textbf{II} & \textbf{III} \\ \hline
    \multicolumn{1}{c|}{1} & \cellcolor[HTML]{FFCCC9}0.1271 & \cellcolor[HTML]{9AFF99}0.0037 & 0.0358 \\
    \multicolumn{1}{c|}{2} & 0.1043 & 0.0133 & \cellcolor[HTML]{FFCCC9}0.0667 \\
    \multicolumn{1}{c|}{3} & \cellcolor[HTML]{9AFF99}0.1034 & 0.0114 & \cellcolor[HTML]{9AFF99}0.0274 \\
    \multicolumn{1}{c|}{4} & 0.1254 & \cellcolor[HTML]{FFCCC9}0.0188 & 0.0344 \\ \hline
    \multicolumn{4}{c}{Yeoh third order} \\
    \multicolumn{1}{c|}{} & \multicolumn{3}{c}{NMSE} \\
    \multicolumn{1}{c|}{\multirow{-2}{*}{Loc.}} & \textbf{I} & \textbf{II} & \textbf{III} \\ \hline
    \multicolumn{1}{c|}{1} & \cellcolor[HTML]{FFCCC9}0.1469 & 0.0164 & 0.0345 \\
    \multicolumn{1}{c|}{2} & 0.0853\cellcolor[HTML]{9AFF99} & 0.0269 & \cellcolor[HTML]{FFCCC9}0.0583 \\
    \multicolumn{1}{c|}{3} & 0.1101 & \cellcolor[HTML]{9AFF99}0.0124 & \cellcolor[HTML]{9AFF99}0.0259 \\
    \multicolumn{1}{c|}{4} & 0.1373 & \cellcolor[HTML]{FFCCC9}0.0531 & 0.0307 \\ \hline
    \multicolumn{4}{c}{Neo-hookean} \\
    \multicolumn{1}{c|}{} & \multicolumn{3}{c}{NMSE} \\
    \multicolumn{1}{c|}{\multirow{-2}{*}{Loc.}} & \textbf{I} & \textbf{II} & \textbf{III} \\ \hline
    \multicolumn{1}{c|}{1} & 0.1027 & \cellcolor[HTML]{9AFF99}0.0031 & 0.0446 \\
    \multicolumn{1}{c|}{2} & 0.1066 & \cellcolor[HTML]{FFCCC9}0.0438 & \cellcolor[HTML]{FFCCC9}0.0628 \\
    \multicolumn{1}{c|}{3} & \cellcolor[HTML]{FFCCC9}0.1301 & 0.0172 & \cellcolor[HTML]{9AFF99}0.0357 \\
    \multicolumn{1}{c|}{4} & \cellcolor[HTML]{9AFF99}0.0959 & 0.0307 & 0.0403 \\ \hline
    \end{tabular}
    \label{tab:results_errors}
\end{table}

Further analysis per point shows that the NMSE for scenarios II and III (single variable fitting) ranges between 0.003 and 0.0667; however, the force or torque response differs from the experimental data, as shown in section 3.2. In contrast, scenario I (dual variable fitting) presents an NMSE range from 0.085 to 0.147, higher than the other scenarios; nevertheless, it ensures a realistic response, as shown in Figs. \ref{fig:ogden_kinet}, \ref{fig:yeoh_kinet}, and \ref{fig:neoh_kinet}.

\subsection{Generalisation of Yeoh model parameters}

\begin{figure}
    \centering
    \includegraphics[scale=1]{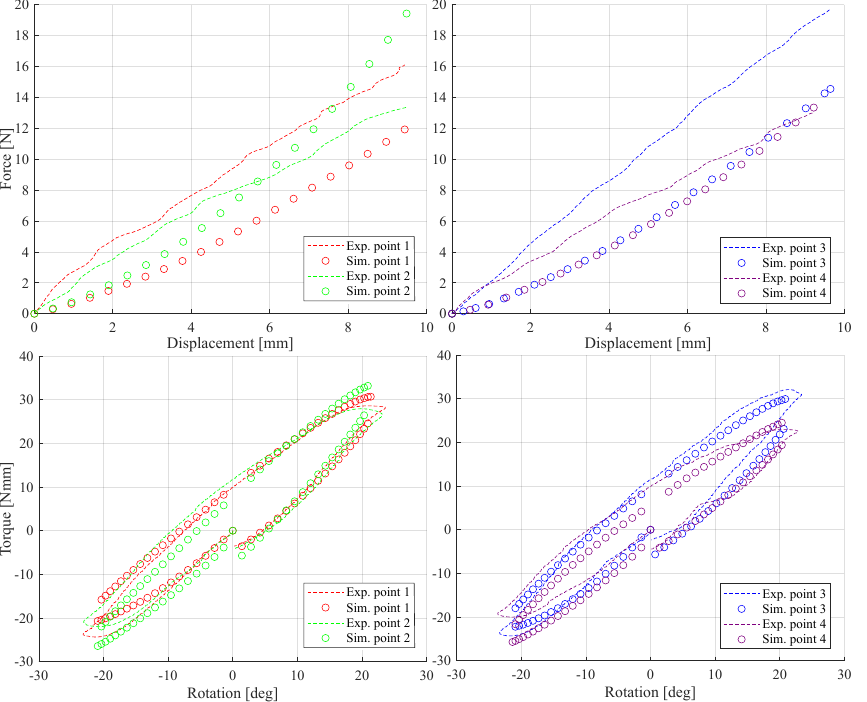}
    \caption{Generalised fitting curves. The red curves shows the best fit for the entire set of characterisation points, and in gray the smoothed experimental data.}
    \label{fig:gener_curves}
\end{figure}

As the Yeoh model showed the lowest NMSE, this material model is used to find a general fitting across the four points. Following Eq. \ref{eq:global_tor-for}, the lowest mean NMSE is 0.1724 $\pm$ 0.0392 for the set of parameters: $c_1=0.0129$, $c_2=-2.016e-3$ and $c_3=2.7623e-4$. The force and torque response fitted differently, as shown in Fig. \ref{fig:gener_curves}. The force curves exhibit a closer fit after 8mm of displacement for points 2-4; conversely, the simulation data are away from the experimental data before 8mm. In contrast to the torque curves, it presents a better fit for all experimental curves along the transitions 0 to -22.5, -22.5 to 22.5, and 22.5 to 0. However, the simulation torques at -22.5 or 22.5 deg are farther from the experimental ones.

\section{Discussion}

The characterisation method in this work demonstrates the importance of incorporating two variables into the curve-fitting process for both force and torque responses. As presented in Section 3.2, the in silico force and torque are closer to the experimental data when both variables are included in the process, in contrast to fitting only one variable (either force or torque), which results in a greater deviation for the variable that was not included. Although this seems intuitive, the main implication of fitting a single variable is that it produces unreliable results for directions different from the fitted variable. Hence, a set of coefficients obtained from a single variable is limited in its ability to predict responses or outcomes in other directions. For instance, Sengeh et al. presented a multi-material model of a transtibial residuum limb that fits normal force-time indentations \cite{Sengeh2016}, however, if this model is used to simulate the interaction with a socket the only reliable results are the ones related to the normal direction. Similarly, the material coefficients found in \cite{Frauziols2016} presents the same limitation as the normal force is used as the fitting curve. In this context, interaction metrics such as the pressure distribution and shear stress may not be accurate to truly understand the interaction because the forces involved are along multiple axis or fully depend on tangential forces.

According to the NMSE along different scenarios and points of analysis, the in-silico force or torque can fit fairly close to the corresponding experimental data, where the NMSE is between 0.0037 and 0.0667. Nevertheless, the same set of coefficients does not provide an accurate response for both variables at the same time. This behaviour can be seen within the force fitting scenario (i.e., purple curves in Fig. \ref{fig:ogden_kinet}, Fig. \ref{fig:yeoh_kinet}, Fig. \ref{fig:neoh_kinet}) where the torque response differs considerably from the torque experimental data. Moreover, this behaviour was also found at different points, varying the geometry and boundary conditions (location of the bones). These implications are identified in the most common material models used to simulate the interaction between a human limb and a cuff or socket. 

As shown in the force and torque responses across the three material models, adding a second variable to the indentation process is crucial to obtain reliable parameters. Oddes et al. included the displacement of certain nodes on the surface close to the indenting region, showing multiple suitable sets of coefficients when only force is included \cite{Oddes2023}. On the one hand, this sensitivity analysis showed the need to include a second variable; similarly, our work also proves the same need of two variables to find a reliable material model. On the other hand, their second variable relied on the exact position of the node, which can increase the complexity of the setup and present practical challenges. Contrary to our methodology, which implements a single indenting setup that provides both variables required to the characterisation method.

The NMSE also varies in different ranges across models and scenarios (i.e., (I) the sum of both NMSE, (II) the minimum NMSE considering only the torque curve, and (III) the minimum NMSE considering only the force curve). The NMSE showed that the force error contributed more than the torque error, which can be seen when scenarios I and III have the same torque curves (Fig. \ref{fig:ogden_kinet} first row). However, these matching scenarios occurred at only one of the four points for all of the material models and not consistently in the same region.

As the phantom soft tissue is represented by a single material, a unique set of parameters is generalised, determining a closer fit for the torque curves rather than the force, as shown in Fig. \ref{fig:gener_curves}. However, the mean of NMSE is still higher than other single-variable characterisation methods where the error varies between 7-11\%, which can be limited by the discrete number of sets given by the parameter sampling.  

\section{Conclusions \& Future Works}

This characterisation method demonstrates the importance of incorporating two variables into the characterisation process at a practical scale for wearable devices, and allows to find a better set of parameters of a material model that interacts and provides normal and shear forces in a realistic way. It also has the potential to create models that can be used as foundation to simulate, at the closest level possible, the cuff and the human limb involved in the physical interaction between the human and wearable robot.

This dual-variable approach enhances the material model’s predictive capability and offers a robust foundation, as the fitting combination of the force and torque curves provides a suitable set of coefficients with an NMSE lower than 0.1034 across multiple material models.

Future works will be focused on reducing the mean NMSE to converge into parameters that resembles the experimental data. In addition, these approach will be also focus on multiple material models to distinguish the skin and soft tissue complex of the human limbs. Finally, the method will also include global optimization algorithms enhancing the parameter search and reduce the error in both variables.

\section*{Author contributions}

\textbf{Felipe Ballen-Moreno}: conceptualisation, methodology, software, validation, formal analysis, investigation, data curation, writing - original draft \& editing, visualization. \textbf{Pasquale Ferrentino}: conceptualisation, methodology, formal analysis,  writing - original draft \& editing. \textbf{Milan Amighi}: software, writing review. \textbf{Bram Vanderborght}: review \& editing and funding acquistion. \textbf{Tom Verstraten}: conceptualisation, formal analysis, funding acquistion, review \& editing, supervision.

\section*{Acknowledgements/Funding sources}
This work was supported by internal fundings from Vrije Universiteit Brussel. Computational resources used in this work were provided by the VSC (Flemish Supercomputer Center), funded by the Research Foundation - Flanders (FWO) and the Flemish Government.

\section*{Declaration of generative AI}
During the preparation of this work, the authors used Copilot to verify grammatical structure, correct typos, and improve the clarity of the text. After using this tool, the authors reviewed and edited the content as needed and take full responsibility for the content of this publication.

\bibliographystyle{elsarticle-num} 
\bibliography{bibliography.bib}

\section*{Supplementary data}

Supplementary data are attached in Research Data item, showing the simulation results of three main material models at four points, figures of the coefficients used in each simulation attempt, the force and torque response.  

\end{document}